\documentclass[letterpaper]{article}
\usepackage{iccc}

\usepackage{flushend} 
\usepackage{times}
\usepackage{helvet}
\usepackage{courier}
\usepackage{graphicx}
\usepackage{natbib}

\title{Generative Design in Minecraft: \\
Chronicle Challenge}
\author{Christoph Salge\\
School of Computer Science\\
University of Hertfordshire\\
Hatfield, United Kingdom\\
ChristophSalge@gmail.com\\
\And
Christian Guckelsberger\\
School of Electronic Engineering and Computer Science\\
Queen Mary, University of London\\
London, United Kingdom\\
christian.guckelsberger@qmul.ac.uk\\
\AND
Michael Cerny Green\\
Game Innovation Lab\\
New York University\\
New York, USA\\
mcg520@nyu.edu\\
\And
Rodrigo Canaan\\
Game Innovation Lab\\
New York University\\
New York, USA\\
rmc602@nyu.edu\\
\And
Julian Togelius\\
Game Innovation Lab\\
New York University\\
New York, USA\\
julian@togelius.com\\
}
\begin{document} 
\maketitle
\begin{abstract}
\begin{quote}
We introduce the Chronicle Challenge as an optional addition to the Settlement Generation Challenge in Minecraft. One of the foci of the overall competition is adaptive procedural content generation (PCG), an arguably under-explored problem in computational creativity. In the base challenge, participants must generate new settlements that respond to and ideally interact with existing content in the world, such as the landscape or climate. The goal is to understand the underlying creative process, and to design better PCG systems. The Chronicle Challenge in particular focuses on the generation of a narrative based on the history of a generated settlement, expressed in natural language. We discuss the unique features of the Chronicle Challenge in comparison to other competitions, clarify the characteristics of a chronicle eligible for submission and describe the evaluation criteria. We furthermore draw on simulation-based approaches in computational storytelling as examples to how this challenge could be approached.
\end{quote}
\end{abstract}

\section{Introduction}

In this paper we introduce the new \textit{Chronicle Challenge} as additional, optional part of the \textit{Generative Design in Minecraft (GDMC) Settlement Generation Competition}\footnote{Further information about the competition can be found on our website: http://gendesignmc.engineering.nyu.edu/}~\citep{Salge:2018}. For the original competition, participants are required to submit code that creates a Minecraft~\citep{game:Minecraft} settlement in an unseen map. 
The goal is to foster interest in the problems of \textit{adaptive} and \textit{holistic} procedural content generation (PCG)~\citep{shaker2016procedural,katecompton2016,short2017procedural}, and to provide a platform on which different solutions can be compared. Rather than starting from a blank slate, an \textit{adaptive} generator must produce an artefact, i.e. a settlement, in response to existing content such as the map layout and climate. Furthermore, by \textit{holistic} we mean that different types and aspects of content should fit well with each other, and potentially echo interactions in-between \citep{antoniosliapis2015,Liapis2019OrchestratingGG}. For instance, a good entry would reflect how a settlement has been constrained and influenced by e.g. mountains and climate, but also how this settlement has shaped the surrounding landscape over time. 

There are numerous examples of well-crafted human settlements that master these challenges, yet no human comparable AI solution exists. Eventually, we want to see generated settlements that are on par with human creations, and understand the underlying creative process. As in many other creative tasks, there is no well-defined, ``optimal'' solution that could be fully captured, or even approximated, by an objective function \citep[cf.][Ch.~8]{smith2012mechanizing}. This property characterizes many challenges in computational creativity (CC), distinguishing the field from general AI research \citep{colton2012computational}. We can thus identify adaptive and holistic settlement generation as a CC challenge.

Given the vague nature of the objectives, the artefacts are judged by human referees based on the criteria of \textit{Adaptivity, Functionality, Evocative Narrative and Aesthetics}. Each criterion is evaluated based on a list of illustrative questions. \textit{Adaptivity} is defined as making a settlement that fits into the given map. In the previous challenge, participants dedicated a lot of work to appropriate building placement, and to generate buildings that reflect the existing natural resources. However, the current approaches feature very little ``big picture'' adaptation, e.g. algorithms do not yet decide whether a large farming village or a fortress would be the more appropriate settlement for a given map. Adaptivity~\citep{lopes2011adaptivity} is one of the core aspects of this challenge, and also permeates the other scoring categories. \textit{Functionality} is defined as providing affordances \citep{gibson1966senses} to hypothetical players and villagers. Here we benefit from the advantage that Minecraft is a game, and as such the world affords specific gameplay-relevant actions to the player \citep{cardona2013cognitivist}. Subsequently, structures in Minecraft can provide additional affordances relevant to player goals. Examples include bridges to allow for extra mobility, houses to protect from monsters, etc. The competition also considers affordances which have no direct relevance in the game, but would in reality. The criterion of \textit{Aesthetics} is less about building a settlement that is beautiful, and more about avoiding errors that are obvious to humans, such as awkward placement or lack of proportion.

The first three criteria have been approached by past GDMC competition entries in a variety of ways \citep[e.g.][]{brightmoore}, but there has yet been little progress on creating an \textit{Evocative Narrative}. The challenge here is to generate a settlement which, as an artefact, implicitly tells a story of how it came about, and of the people that inhabit it. Real-world settlements tell such stories, but also human-build settlements in Minecraft; people express cultural influences, and their settlements have an imagined or actual history that brought it about in a casual way. A settlement is the transient outcome of a creative process that usually involves many agents, and (computational) creativity researchers have long agreed that creativity does not happen in a vacuum \citep{jordanous2016four}. However, procedurally generated settlements often lack an Evocative Narrative.

To advance this aspect of our challenge we have therefore decided to introduce an optional bonus challenge, the Chronicle Competition, which adds the task of generating an explicit narrative, captured in natural language text. In the remainder of the paper, we outline the competition and discuss which unique challenges it offers in comparison to other benchmarks. To give participants a head start, we furthermore discuss how existing generative approaches could be applied to the challenge. We particularly consider approaches in \textit{computational storytelling}, a CC subfield concerned with the study of algorithms capable of generating fictional narratives \citep{gervas2009computational, berov2018character}. We thus connect this competition to existing work in CC, and yet open it up to researchers and the general public with an interest in PCG more generally.

\begin{figure*}
    \centering
    \includegraphics[width=0.49\textwidth]{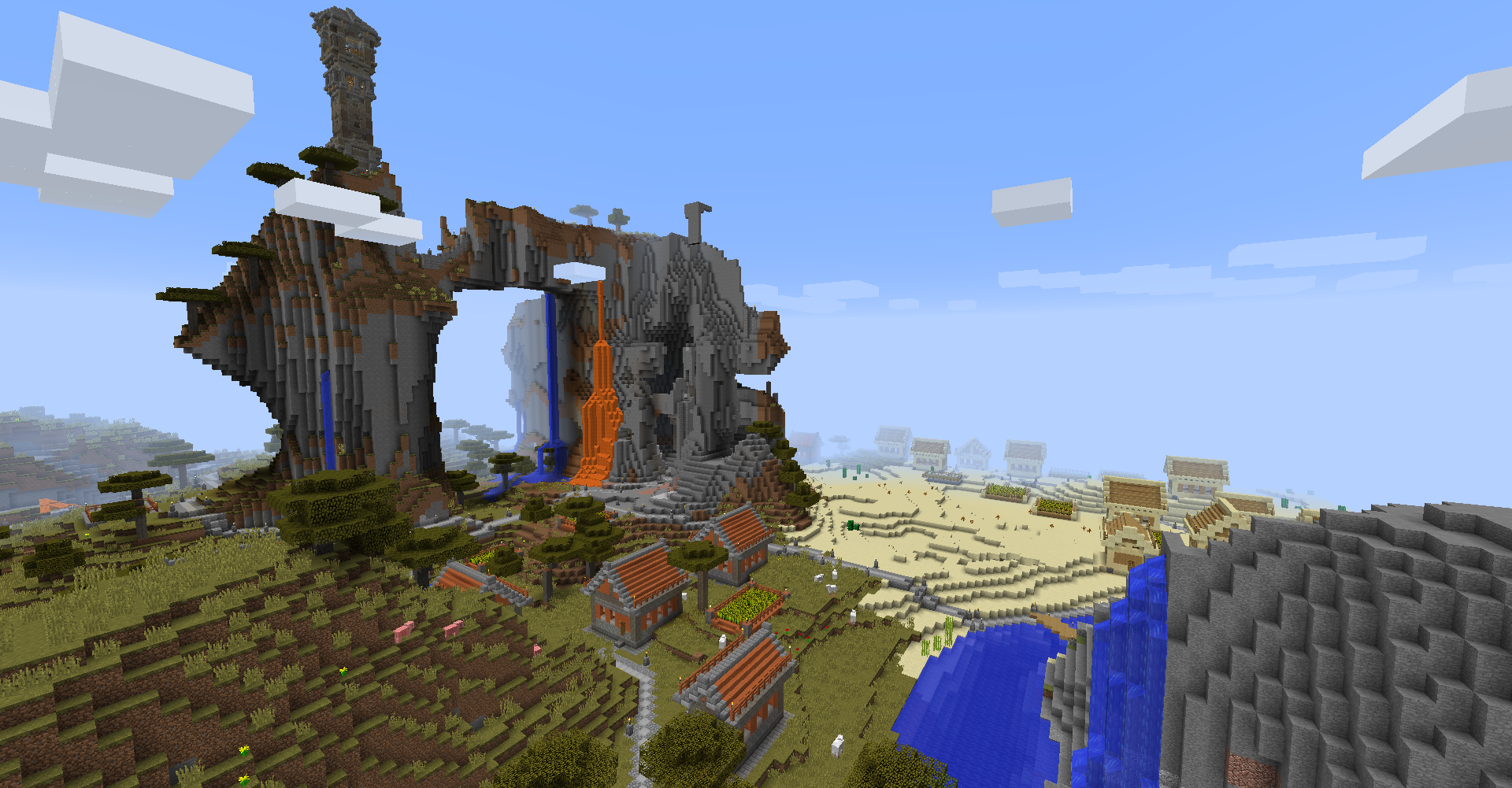}
    \includegraphics[width=0.49\textwidth]{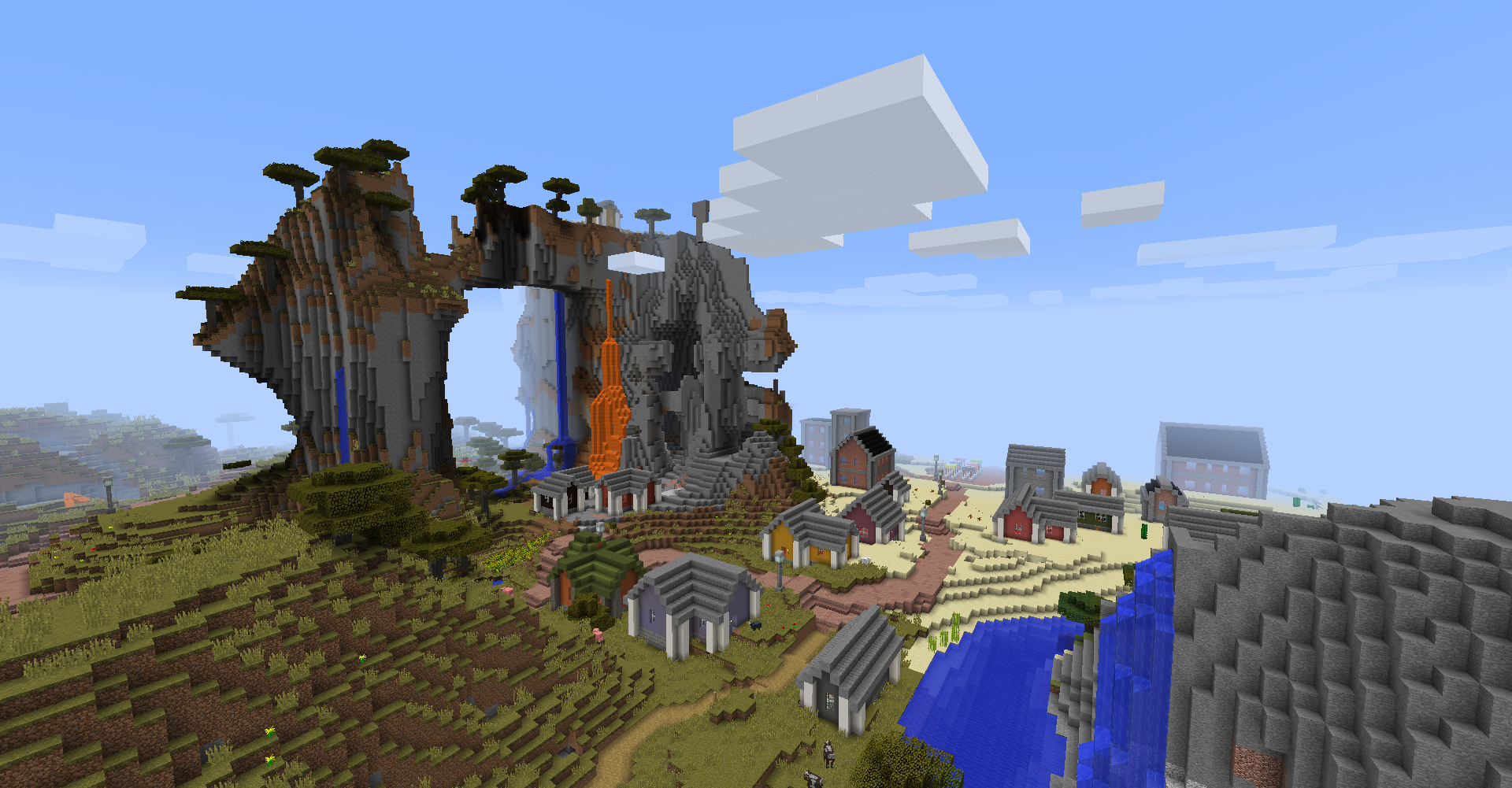}
    \caption{Two examples of settlements from the 2nd GDMC competition, produced by different generators on the same map. To illustrate the idea of \textit{fit} consider the following two chronicle fragements:\\ 1) \textit{... we settled next to the desert tribe and placed a watchtower to keep an eye on them ...}\\ 2) 
    \textit{... our village once was a trading post - with the rich traders' big houses cluttered around the central market square ... }\\For a high ``Fit'' score it should be evident which fragment belongs to which settlement.} 
    \label{fig:my_label}
\end{figure*}





\section{The Chronicle Competition}



The main task for the competition is to generate a \textit{chronicle}, i.e. a written text about the history of a Minecraft settlement, and place it inside that settlement as a Minecraft book. We are deliberately vague about what exactly a chronicle is; to illustrate the range of encouraged submissions, we provide a number of examples. A chronicle could be a text written by different people during different times in the development of a settlement, or it could be a retelling of the town's history from a single, modern perspective, or even a tourist guide to historic buildings and places in the city. It can be written in different styles, and focus on different aspects, such as the lives of certain people, or buildings, or communities, etc. The chronicle can feature unreliable narrators and contradicting viewpoints. We explicitly encourage the use of focalization \citep{gervas2009computational}, i.e. the restriction of what is being told to what might have perceived by somebody in the scene. These examples are not exhaustive, and we encourage a wide variety of different submissions to delineate the scope further in the future. At this point, the only hard requirements are that the submission is in English and relates to specific generated settlements and how they came about. 

Entries will be evaluated in terms of their (i) \textit{Overall Quality}, and (ii) their \textit{Fit} for a given settlement. For the evaluation of \textit{Overall Quality}, we rely on the idea of producing objectivity by inter-subjectivity;  each text will be evaluated by a number of human judges with diverse views. This is quite customary in competitions where humans or AIs generate creative game artefacts  \citep{shaker20112010,stephenson20182017,khalifa2016general,khalifa2017general}. It alleviates the problem that there are no commonly accepted, computational measures for narrative quality. For example, some narratologists argue that a \textit{story} is different to a \textit{plot} or \textit{narrative}, in that the earlier only represents a list of events, but the latter connects those with causal relations. \citet{labov1972language} defines a minimal narrative as two states and a transition or movement in-between, where such a transition could be given by a causal relationship. However, others might consider causality as an aid in understanding a story, rather than as a strict requirement of a narrative \citep[cf.][]{gervas2009computational}. While there are some interesting metrics \citep[e.g.][]{berov2017towards}, automatic evaluation comes with the additional problem of elevating one or several specific metrics which then becomes the sole goals of optimization. Instead, we consequently encourage participants to use such metrics for the evaluation of their created artefacts, e.g. in ``generate-and-test'' approaches \citep[cf.][]{togelius2011search}, but have human judges evaluate how well those criteria work. Since what makes a good narrative is still debated, we employ a range of soft constraints that are enforced through scoring. Rather than disqualifying a submission based on e.g. the absence of causality, we want to leave it to our judges to potentially give a lower score, but maybe also reward other, good features of the chronicle. The competition thus supports the discovery of new elements that make good narratives, and of evaluation criteria that could prove useful in computational storytelling, narratology and related fields. Our aim is to establish a relatively low hurdle for a minimally sufficient solution to encourage participation, but to also have a lot of room for improvement \citep{togelius2016run}.

The second criteria, \textit{Fit}, will also be judged by humans, but is subject to more specific instructions. ``Fit'' is about how well a text corresponds to a specific settlement. Given that we usually have three competitions maps, imagine that a generator produces a settlement for each of those maps, and a chronicle for each settlement. Now imagine that we shuffle those chronicles around; would you still be able to assign each chronicle to the settlement it was originally generated for? Entries that show a clear relationship between the settlement and text would score high on Fit. This criterion inherits the focus of the overarching competition on adaptive PCG, for which generated content must be responsive to some other existing content. Ultimately, this touches upon an old problem of text generation, namely how to produce a text that is genuinely \textit{about} something \citep{Woods81}. We think that it is possible to approach this without delving into the deep philosophical issues arising here, but nevertheless those issues are relevant. We believe that Fit has been neglected in computational storytelling so far, because existing systems mostly feature only one story world of limited complexity. Consequently, the generated narratives naturally fit into the bigger picture. Yet, we consider this an interesting challenge to inspire further developments in the domain, in particular for modular, service-based systems \citep{leon2011stella, veale2013service, gervas2017deconstructing} capable of generating narratives from exchangeable story worlds. In the Chronicle Competition, generators are evaluated on different maps, based on settlements with arbitrary complexity.

\section{Possible Approaches}

In this section we discuss possible approaches to chronicle generation based on existing work in computational storytelling, but we want to stress that there is no restriction in techniques for this competition; we explicitly want to encourage both \textit{amateurs} and \textit{specialized researchers} to find \textit{new} solutions to the challenges involved. We highlight possible starting points, but also point out the challenges of the individual approaches for which this competition would offer an interesting and comparable benchmark. While there is a range of computational storytelling techniques, we focus mostly on those that can in one way or another deal with the holistic adaptation to existing content. 

Recent data-driven approaches based on neural networks are capable of producing descriptions for photos \citep{donahue2015long}, or even to synthesize 3D scenes based on textual descriptions \citep{reed2016generative}. The first technique could e.g. be used to generate data for storytelling from the perspective of the player character wandering through an existing settlement. The second technique could ultimately allow to inversely generate a settlement from an existing chronicle. Unfortunately, these approaches usually require a lot of training data, which in the case of chronicles for Minecraft settlements is not available. It is also unclear if they would scale to the complexity required to tell a story about a whole settlement. However, they might be useful in modular form, for example to generate the individual buildings of a settlement from a text description, to insert descriptions of buildings or natural sights into the chronicle, or to identify interesting elements in a settlement. For example, there is already a model that can generate design descriptions for single buildings in Minecraft~\citep{AIIDE1818112}. Existing approaches to generating text in specific styles could also be of interest, but they yet often struggle with adhering to a cohesive structure.

There are also a wide range of more structural approaches to computational narrative generation. Many existing approaches have been surveyed by \citet{gervas2017deconstructing} and \citet{kybartas2017survey}, and can be split into roughly two categories. \textit{Simulation-based approaches} rely on simulating the interaction of agents and turning the recorded events into a narrative \citep[e.g.][]{theune2003virtual, leon2011stella, berov2018character, y2015computer}. A game-based example is the history generation of Dwarf Fortress \citep{dwarffortress,hall_2014}, where generations of characters are born, fight, and die, producing a logbook of many events. For our competition, we might imagine some algorithm that successively builds a settlement and records events such as newly built houses, removing forests, etc. The problem with this naive approach though is that it misses out on establishing causal relationships between the events, and some might thus consider the resulting artefact only a basis for, but not an actual narrative. One popular means to overcome this in the cited work is to specify the beliefs and desires of the involved agents as source of meaningful, causal interaction.

The second category is given by \textit{planning-based approaches} \citep[e.g.][]{riedl2010narrative} leveraging propositional logic to generate narratives. Agents can be modelled with specific goals, and an ontology can be defined that describes how certain actions produce certain outcomes. The planning agents then perform actions leading to a certain goal, which allows for a descriptions with a causal structure. I.e. a settlement's goal might be to have food production, and building farms might provide food production. In textual form this might lead to: \textit{``We built farms on the slopes of the mountain to feed our people''}. The difficulty here is to design such an ontology in the first place which fits well into the world, but this should generally be possible in a game such as Minecraft. With such an ontology, the planner could produce the narrative structure and simultaneously be used to plan and build the settlement. It might then be worthwhile to add some noise to get a more exciting narrative. For instance, part of the settlement could burn down, an event that rational agents would not trigger as part of their plan. 

Another issue with this kind of approach is to define the right kind of agent, with believable and interesting goals. The focal point of stories are mostly people, and their desires are relatable to us. A story about overcoming starvation or dangers is interesting, a story about an agent that wants to build 15 houses and then builds 15 houses is less so. Here the chronicle format might be a bit add odds with most projects in computational storytelling which are very character focused. However, it is important to note that similarly, a settlement is shaped and experienced by characters; a city's population can be modelled as a single or multi-agent system with relatable human goals, such as, we felt threatened so we decided to build walls to protect us. Exemplary figures, such as the mayor, can serve as character and embody those views to give them a human perspective. Similarly, a multi-agent approach could simulate the interaction of different factions, each with their own goals, leading to conflicting actions. A trading guild might want to build a harbour, but the local farmers might sabotage this project because they fear competition with their crops, etc. 
Finally, a lot of these approaches can be brought together. Given that participants can design both the settlement generator and the chronicle generator one approach would be to first design a process that simulates the causal chain of event that brings about a settlement. This can be done with a variety of simulated characters, ideally driven by believable motivations and encountering interesting random events. This, in essence, is very similar to what a lot of story generators do already. Then, this has to be followed by designing two projection functions, that translate this process a) into a textual history and b) into a 3d representation of the settlements. Both the settlement and the chronicle can be seen as imperfect projections, capturing different details, or a much richer actual history. Again, this is a problem not uncommon in computational storytelling, where several generators have a story graph that then gets translated into a text. 

While we wanted to illustrate the breadth of existing approaches, we also want to stress that a minimal solution to producing a chronicle could work with very simple techniques, for example a text where placeholders are filled based on parameters derived from a settlement, such as \textit{``We build a city in THE DESERT''}. But at the same time, the framework and challenges outlined here can be tackled with a lot of different, sophisticated methods and it would be interesting to see, if relying on them produces noticeable better results. We think that this challenge could serve as a platform to compare different approaches in a common task.

\section{Future Plans}

The Chronicle Challenge has once already been part of the annual GDMC competition. At present, both challenges are interwoven, i.e. participants interested in chronicle generation must also provide a settlement generator, but not vice-versa. This allows for more freedom in the chronicle generation: a chronicle could be written post-hoc after the generation of a settlement based on the final artefact only, or alternatively as a settlement unfolds, leveraging a tight and unrestricted communication with the settlement generator, and potentially even interacting with it. The downside of this is that participants who are mainly interested in chronicle generation must also deal with the more general PCG challenges of settlement generation. As a consequence, the quality of chronicles may heavily depend on the quality of the settlement generator, and cannot be judged independently.

For future competitions we thus consider to offer a \textit{standalone} Chronicle Challenge to attract more researchers from specialized fields such as computational storytelling. One option would be to ask participants to provide a chronicle generator for one specific Minecraft map with an existing settlement. However, this would be quite challenging in terms of extracting information about the settlement from a block-based representation. We may provide additional information such as building labels or historic events alongside the actual map, but this would require to first figure out what important information from settlement generation must be preserved for good chronicles. As an alternative option, we may give all participants one generator that creates a settlement over time and offers rich information along the way. All participants would thus have access to the same, rich time-sensitive data as input to their chronicle generator. 

For now, our plans are to rerun the chronicle competition as is and point interested participants to existing, open-source entries from previous years, that could be extended for chronicle generation. We hope that this introduction excited prospective participants, and gives them a head start in the competition.

\section{Acknowledgments}
CS is funded by the EU Horizon 2020 programme / Marie Sklodowska- Curie grant 705643. CG is supported by EPSRC grant EP/L015846/1 (IGGI). RC gratefully acknowledges the financial support from Honda Research Institute Europe (HRI-EU). Many thanks to Leonid Berov and Dino Pozder for feedback on the initial idea for this competition through the lens of computational storytelling and narratology as part of the 2019 Dagstuhl seminar ``Computational Creativity Meets Digital Literary Studies'' (19172).






\bibliographystyle{iccc}
\bibliography{iccc}

\end{document}